\documentclass{article}

\usepackage{PRIMEarxiv}

\usepackage[utf8]{inputenc} % allow utf-8 input
\usepackage[T1]{fontenc}    % use 8-bit T1 fonts
\usepackage{url}            % simple URL typesetting
\usepackage{booktabs}       % professional-quality tables
\usepackage{array}
\usepackage{amsmath}        % math environments and \dfrac
\usepackage{amsfonts}       % blackboard math symbols
\usepackage{nicefrac}       % compact symbols for 1/2, etc.
\usepackage{microtype}      % microtypography
\usepackage{fancyhdr}       % header
\usepackage{graphicx}       % graphics
\usepackage[font=small,labelfont=bf,hypcap=true]{caption}  % smaller captions, bold label, link to float top
\usepackage{multirow}
\usepackage{xcolor}
\graphicspath{{assets/}}
\usepackage{tabularx}
\usepackage{charter}                      % Charter serif (closest free match to Tiempos)
\usepackage[scaled=0.92]{helvet}          % Helvetica for sans-serif metadata elements

\makeatletter
\renewcommand{\@maketitle}{%
  \vbox{%
    \hsize\textwidth
    \linewidth\hsize
    \vskip 0.1in
    {\raggedright\fontsize{17}{20}\selectfont\bfseries\hyphenpenalty=10000\exhyphenpenalty=10000 Stream-Based Active Learning with Cooperative Neural Networks for Data-Efficient Partial Inverse Design: An Automotive Glass Run Channel Case Study\par}
    \vskip 0.2in
    {\normalsize\textbf{Agung Nugraha\,$^{1}$, Hyerin Kwon\,$^{2}$, Heungjun Im\,$^{1}$, Gian Antariksa\,$^{1}$, Jihwan Lee\,$^{1*}$}\par}
    \vskip 0.1in
    {\small
    $^{1}$\,Department of Industrial and Data Engineering, Major in Industrial Data Science and Engineering, Pukyong National University, Busan, 48513, Republic of Korea\\
    $^{2}$\,DRB Co., Ltd., Busan, 46329, Republic of Korea\par}
    \vskip 0.4in
  }
  \footnotetext{$^{*}$Correspondence: jihwan@pknu.ac.kr}
}
\makeatother

\title{}
\author{}

\begin{document}
\maketitle

\begin{abstract}
Inverse design in engineering often runs into a simple problem. Each labeled training sample must be produced through expensive simulation, so building a large dataset is slow and costly. This study addresses that problem for partial inverse design, where only some design variables are specified and the rest must be inferred to reach a target performance value. We propose CoNN-AL, a framework for data-efficient partial inverse design that adds stream-based active learning to the Cooperative Neural Network with Denoising Autoencoder (CoNN-DAE). The model estimates predictive uncertainty through Monte Carlo dropout and uses it to decide, in real time, which incoming candidate samples are worth labeling, so the limited labeling budget is spent on the most informative designs. We validate the framework on a real-world automotive glass run channel dataset of more than 900,000 unique simulated designs. With only 20,000 actively selected labels, about 2.3\% of the training pool, CoNN-AL reaches R-squared values of 0.967 to 0.982 across all missing-variable levels, approaching the upper-bound models trained on far more data. It reaches R-squared of at least 0.95 with 30 to 40\% fewer labels than random sampling at the more difficult missing-variable levels and, at the most challenging level, is the only strategy in this study to reach R-squared of 0.98. Together with this work, we publicly release the dataset to support future research on data-driven design.

\end{abstract}


\section{Introduction}
\label{sec:introduction}

Conventional engineering design has long relied on iterative prototyping and repeated forward analyses, particularly finite element analysis (FEA), to determine whether candidate geometries or material configurations satisfy prescribed functional or mechanical requirements. However, this trial-and-evaluation paradigm is often computationally intensive, time-consuming, and dependent on substantial computing resources for repeated FEA simulations, especially when the design space is large, nonlinear, and high-dimensional~\cite{Ebrahimian2024_JU83KRZF, Dabiri2019_465DBYGX}. In industrial component design, numerical analysis must often be completed within a practical time frame to support repeated design iterations~\cite{Dabiri2019_465DBYGX}. As a result, data-driven machine learning approaches have increasingly been investigated as efficient alternatives or surrogates for conventional simulation-based analysis such as FEA and CFD, with the aim of approximating simulation outcomes at substantially lower computational cost~\cite{Madani2019_A6T8F5I3, Ebrahimian2024_JU83KRZF, Dabiri2019_465DBYGX, Sun2019_28NHRMVP, Wang2022_RTU4KDVZ, Vlah2022_JWG5JYN4}. Building upon this development, data-driven inverse design has emerged as a promising paradigm in which machine learning models are used to infer candidate design-variable configurations from prescribed performance targets, thereby reducing reliance on repeated forward simulation loops during design exploration~\cite{Jiang2022_KXSGVNZG, Brzin2024_6ZKNZIMH, nugraha_cooperative_2025, Li2026_8FEFCHZT, Ha2023_C4XI4I2D}.

In practice, inverse design problems are often further complicated by settings in which only a subset of design variables can be freely determined, while the remaining variables are fixed in advance by manufacturing requirements, operational conditions, material availability, or regulatory constraints. This setting closely relates to constrained optimization problems, which seek feasible or optimal solutions under prescribed variable restrictions and design conditions~\cite{Mathern2021_XZ7IKVLB}. Bayesian optimization has been widely adopted in prior studies to address such constrained and computationally expensive problems~\cite{Amini2025_JNACVRUT, Wang2024_94QLV7FR}. However, Bayesian approaches can become computationally demanding and sensitive, particularly when the objective function is complex, the design space is high-dimensional, or the dataset is large~\cite{Amini2025_JNACVRUT, Mathern2021_XZ7IKVLB}. In our previous work, we formulated this challenge as a partial inverse design problem and proposed a Cooperative Neural Network (CoNN) framework, which integrates a surrogate model and an imputation model through a cooperative learning mechanism to infer undetermined variables while satisfying both the observed constraints and the desired target output. Once trained, CoNN can produce feasible design solutions through a single forward pass, enabling rapid inference without the need for retraining when constraint settings vary~\cite{nugraha_cooperative_2025}. Although this framework effectively addressed the constraint-aware aspect of partial inverse design, it did not explicitly consider another important practical bottleneck related to the high cost of obtaining new labeled samples. In many engineering domains, each additional label requires expensive finite element simulation, laboratory experimentation, or other high-fidelity evaluation procedures~\cite{Ebrahimian2024_JU83KRZF, Madani2019_A6T8F5I3, Mathern2021_XZ7IKVLB}. As a result, even if the inverse design framework itself is efficient at inference time, its practical use may still be constrained by the cost of constructing sufficiently informative training data. This leaves an important gap in partial inverse design research, since beyond handling constrained variable completion, there is also a need to reduce the number of costly evaluations required to achieve strong design performance.

More broadly, data-driven inverse design methods, including neural network-based approaches, typically require large volumes of labeled training data to achieve reliable generalization across the design space~\cite{Lee2023_I94WMQTR, Lu2022_6NWUC4KN}. Given the high cost of obtaining each labeled sample, as discussed above, this creates a fundamental data bottleneck in which the model demands extensive labeled examples to learn complex input-output relationships, yet the cost of acquiring each label severely limits the amount of data that can be practically obtained within a given project budget. A common strategy to construct training datasets is random sampling from the design space, yet this approach treats candidate samples largely without regard to their informativeness, even though some regions may already be well represented while others remain underrepresented and potentially more valuable for model improvement~\cite{Settles2009_3BU3Z6M2, Cacciarelli2024_JUG4J8J8}. Consequently, a substantial portion of the labeling budget may be spent on redundant samples. This inefficiency becomes particularly pronounced in high-dimensional design spaces, where achieving adequate coverage requires exponentially more samples, a phenomenon commonly referred to as the curse of dimensionality~\cite{Hastie2009_J873EIAZ}. Therefore, the practical deployment of data-driven inverse design frameworks in industrial settings is often limited not by the capacity of the model itself, but by the feasibility of constructing a training set that is both sufficiently large and sufficiently informative under a constrained simulation budget.

To address this data bottleneck, active learning has been proposed as a principled framework that enables the model to selectively query the samples expected to be most informative for improving performance~\cite{Settles2009_3BU3Z6M2, Cacciarelli2024_JUG4J8J8, Shang2026_D2NT85FI, Grbcic2025_AIDN7LQG, Han2025_L84YDM2W, Chen2020_N7IL76LC}. Rather than passively receiving labels for arbitrarily chosen data points, an active learner evaluates the potential value of each candidate and requests labels only when the expected benefit justifies the labeling cost~\cite{Settles2009_3BU3Z6M2, Han2025_L84YDM2W, Chen2020_N7IL76LC}. Among the various active learning scenarios, stream-based active learning is particularly well suited for engineering design applications. In this setting, unlabeled candidate samples arrive sequentially, and the model must decide in real time whether to query the oracle for a label or discard the sample~\cite{Settles2009_3BU3Z6M2, Cacciarelli2024_JUG4J8J8, Rozanec2022_CSEXYCTJ}. This is a natural fit for simulation-driven design workflows, where new candidate configurations can be generated continuously and each evaluation by the oracle, typically a finite element simulation or other high-fidelity analysis, is expensive and time-consuming. A common approach to guide the query decision in stream-based active learning is uncertainty sampling, in which the model estimates its predictive uncertainty for each incoming sample and selects those for which it is least confident~\cite{Settles2009_3BU3Z6M2}. By concentrating the labeling budget on regions of the design space where the model is most uncertain, this strategy can achieve comparable or superior predictive performance with significantly fewer labeled samples than random sampling~\cite{Shang2026_D2NT85FI, Hoffmann2025_RTKUPJU2}. Beyond this specific setting, active learning has shown promise in various scientific and engineering domains, including biotechnology~\cite{Kumar2021_UZKKRSM7}, metallurgy~\cite{Shang2026_D2NT85FI}, and materials and structural design~\cite{Tran2020_4FN63ASY, Han2025_L84YDM2W, Grbcic2025_AIDN7LQG, Chen2020_N7IL76LC, Shen2026_YQUWWEMZ, Hoffmann2025_RTKUPJU2}, where it has been used to improve model performance or reduce the number of expensive experiments and simulations required for data acquisition.

Although both partial inverse design and active learning have seen substantial progress, a gap remains at their intersection. On the partial inverse design side, the Cooperative Neural Network enables constraint-aware inference under predetermined design variables, yet it assumes a sufficiently large labeled dataset and does not provide any mechanism for reducing the cost of acquiring new labeled samples. On the active learning side, prior studies that combine active learning with inverse design operate predominantly in pool-based settings, in which the learner selects samples from a fixed candidate set prepared in advance rather than from a continuous stream of incoming candidates. These studies also assume that all design variables are freely selectable during the query process, which makes them incompatible with partial inverse design problems where a subset of variables is predetermined by manufacturing or regulatory constraints. Consequently, to the best of our knowledge, no prior work has integrated stream-based active learning with a cooperative imputation-surrogate architecture, and uncertainty-driven sample acquisition has not been explored as a means to reduce the labeling cost of partial inverse design under realistic engineering design settings. This gap motivates the present study, which aims to extend cooperative neural networks with a stream-based active learning strategy tailored for constrained and expensive partial inverse design problems.

To address this gap, we propose CoNN-AL, a framework that integrates stream-based active learning into CoNN~\cite{nugraha_cooperative_2025}, a cooperative surrogate-imputation architecture for partial inverse design. Predictive uncertainty is estimated from the cooperative surrogate using Monte Carlo dropout and used as an acquisition signal to decide in real time whether each incoming candidate sample should be queried from the oracle or discarded, thereby concentrating the labeling budget on the most informative regions of the design space. We validate CoNN-AL on a real-world automotive glass run channel (GRC) dataset developed in collaboration with a Korean automotive company. A GRC is a sealing component installed along the vehicle window frame whose sealing force must satisfy prescribed target values under fixed geometric and material design constraints. The dataset contains more than 900,000 samples generated by executing a full-factorial parametric design sweep in CATIA~\cite{Mitrache2011_Z5VDMMMF} and evaluating each configuration through MSC Marc finite element analysis~\cite{Javanbakht2017_YINUMPL8}, providing a realistic testbed that reflects the partial variable setting and expensive evaluation typical of industrial engineering design. With this framework and testbed in place, the main contributions of this study are fourfold. First, we propose CoNN-AL, the first integration of stream-based active learning with a cooperative surrogate-imputation architecture for partial inverse design under expensive evaluation. Second, we design an uncertainty-based acquisition strategy for stream-based active learning in inverse design, using Monte Carlo dropout over the cooperative surrogate to estimate predictive uncertainty and guide real-time query-or-discard decisions. Third, we conduct a comprehensive empirical evaluation on a real-world GRC dataset of more than 900,000 samples and show that CoNN-AL achieves substantial label efficiency gains over random sampling under constrained labeling budgets. Fourth, together with this study, we publicly release the GRC dataset as a large-scale simulation-based benchmark to support reproducibility and future research on data-driven inverse design.

The remainder of this paper is organized as follows. Section~\ref{sec:methodology} presents the proposed CoNN-AL framework, including the cooperative surrogate-imputation architecture and the stream-based active learning strategy driven by Monte Carlo dropout uncertainty. Section~\ref{sec:experiment} describes the GRC dataset, the partial inverse design problem formulation, the training phases, and the evaluation metrics used in the experiments. Section~\ref{sec:results} presents and discusses the experimental results, comparing the proposed active learning strategy against an upper-bound reference and a random sampling baseline. Finally, Section~\ref{sec:conclusion} concludes the paper and outlines limitations and directions for future work.

\section{Methodology}
\label{sec:methodology}

This study combines two complementary components to address the partial inverse design problem under limited labeling budgets. The first component is the Cooperative Neural Network with Denoising Autoencoder (CoNN-DAE), a two-model architecture that jointly reconstructs missing design variables and predicts the target performance criterion. The second component is a stream-based active learning framework that selectively acquires the most informative samples for labeling, guided by predictive uncertainty. Together, these two components form the proposed CoNN-AL framework. The CoNN-DAE architecture is illustrated in Fig.~\ref{fig:conn_architecture}, and the active learning loop is shown in Fig.~\ref{fig:al_loop}.

\subsection{Cooperative Neural Network (CoNN-DAE)}
\label{sec:conn}

\begin{figure}[htbp]
    \centering
    \includegraphics[width=1.0\textwidth]{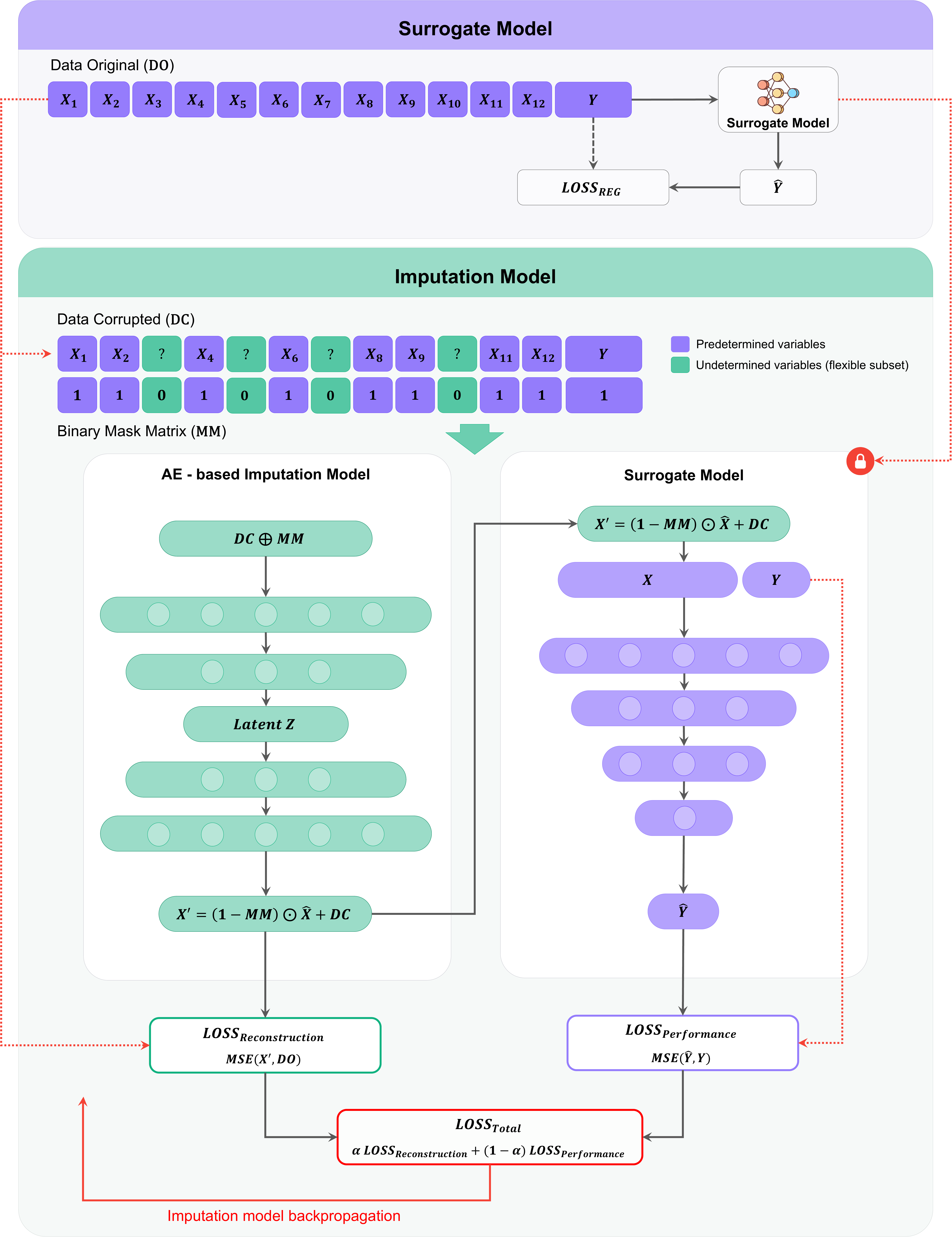}
    \caption{Architecture of the Cooperative Neural Network with Denoising Autoencoder (CoNN-DAE), adapted from Nugraha et al.~\cite{nugraha_cooperative_2025}. The surrogate model (top) maps complete designs to the performance output. The imputation model (bottom) reconstructs missing variables via an autoencoder, jointly optimized through reconstruction and performance losses.}
    \label{fig:conn_architecture}
\end{figure}

This study adopts the Cooperative Neural Network with Denoising Autoencoder (CoNN-DAE) introduced by Nugraha et al.~\cite{nugraha_cooperative_2025} as the base model for partial inverse design. The architecture, illustrated in Fig.~\ref{fig:conn_architecture}, consists of two interconnected networks, namely a regression-based surrogate that maps complete design configurations to the target performance output, and an autoencoder-based imputation network that recovers missing design variables from incomplete inputs. These two networks are trained simultaneously, with the surrogate supplying a performance signal that steers the imputation network toward reconstructions that not only resemble the original design geometrically but also yield accurate performance predictions.

The surrogate network serves as a forward predictor. Given a matrix of complete design variables $\mathbf{X} \in \mathbb{R}^{n \times d}$ and the associated performance vector $\mathbf{y} \in \mathbb{R}^n$ (sealing force in the GRC application), the network learns a function $f(\cdot)$ such that
\begin{equation}
    \hat{\mathbf{y}} = f(\mathbf{X})
    \label{eq:surrogate}
\end{equation}
The training objective is the standard regression loss over $N$ training samples, defined as
\begin{equation}
    \mathcal{L}_{\mathrm{REG}} = \frac{1}{N} \sum_{i=1}^{N} (y_i - \hat{y}_i)^2
    \label{eq:loss_reg}
\end{equation}
This pre-trained surrogate is subsequently frozen and reused within the cooperative loop, where it evaluates designs completed by the imputation network and provides gradient-based feedback on their expected performance.

The imputation network addresses the core task of reconstructing unknown design variables. It receives three inputs, namely the original complete record $\mathbf{DO}$ (used only during training as the reconstruction target), a corrupted copy $\mathbf{DC}$ in which selected variables have been zeroed out, and a binary mask $\mathbf{MM}$ that flags which entries are observed (1) versus missing (0). The mask and corrupted data are concatenated column-wise ($\mathbf{DC} \oplus \mathbf{MM}$) so that the network can distinguish genuinely zero-valued features from masked ones. An encoder $f_{\mathrm{enc}}$ compresses this augmented input into a latent code, computed as
\begin{equation}
    \mathbf{z} = f_{\mathrm{enc}}(\mathbf{DC} \oplus \mathbf{MM})
    \label{eq:encoder}
\end{equation}
A decoder $g_{\mathrm{dec}}$ then maps the latent code back to the full variable space. Because the reconstructed values must remain within physically meaningful bounds, each output dimension is clipped to the observed feature range in the training data~\cite{Nugraha2026_89KTUMZM}, expressed as
\begin{equation}
    \hat{\mathbf{X}} = \min\left(\max\left(g_{\mathrm{dec}}(\mathbf{z}),\, \mathbf{X}_{\min}\right),\, \mathbf{X}_{\max}\right)
    \label{eq:decoder}
\end{equation}
Finally, the known (unmasked) entries from $\mathbf{DC}$ are restored so that only the missing positions are filled by the network, yielding
\begin{equation}
    \mathbf{X}' = (1 - \mathbf{MM}) \odot \hat{\mathbf{X}} + \mathbf{DC}
    \label{eq:merge}
\end{equation}
where $\odot$ is element-wise multiplication. The resulting $\mathbf{X}'$ represents the completed design, ready for performance evaluation by the surrogate.

The key aspect of the CoNN-DAE framework is the joint optimization that couples both networks. After the imputation network produces $\mathbf{X}'$, the surrogate evaluates it to obtain $\hat{\mathbf{y}} = f(\mathbf{X}')$. Two loss terms are computed, namely a reconstruction loss $\mathcal{L}_1$ that penalizes deviations between the completed design $\mathbf{X}'$ and the original design $\mathbf{DO}$, and a performance loss $\mathcal{L}_2$ that penalizes deviations between the surrogate prediction $\hat{\mathbf{y}}$ and the true output $\mathbf{y}$, written as
\begin{equation}
    \mathcal{L}_1 = \mathrm{MSE}(\mathbf{X}',\, \mathbf{DO}), \qquad
    \mathcal{L}_2 = \mathrm{MSE}(\hat{\mathbf{y}},\, \mathbf{y})
    \label{eq:losses}
\end{equation}
A weighted combination forms the total training objective for the imputation network, given by
\begin{equation}
    \mathcal{L}_{\mathrm{TOTAL}} = \alpha \, \mathcal{L}_1 + (1 - \alpha) \, \mathcal{L}_2
    \label{eq:total_loss}
\end{equation}
The coefficient $\alpha \in [0, 1]$ balances geometric accuracy against performance fidelity. During backpropagation, gradients from $\mathcal{L}_2$ flow through the frozen surrogate back into the imputation network, effectively teaching it that a good reconstruction is one whose predicted sealing force matches the ground truth, not merely one that stays numerically close to the original variables. This performance-aware training signal is what separates the cooperative architecture from a standard denoising autoencoder and is the property that makes CoNN-DAE particularly suitable for integration with active learning, where the surrogate's uncertainty estimates directly inform sample acquisition decisions.

Moreover, the cooperative architecture is not limited to a denoising autoencoder. As demonstrated by Nugraha et al.~\cite{nugraha_cooperative_2025}, the AE-based imputation model can be replaced with alternative autoencoder variants. A standard autoencoder is typically deterministic, whereas a Variational Autoencoder (VAE) by Kingma and Welling~\cite{Kingma2013_JR36RT29} is stochastic and probabilistic. A Wasserstein Autoencoder (WAE) by Tolstikhin et al.~\cite{Tolstikhin2017_U9IG7ANN} can be implemented with either a deterministic or a stochastic encoder, depending on the formulation.

\subsection{Stream-Based Active Learning}
\label{sec:active_learning}

\begin{figure}[htbp]
    \centering
    \includegraphics[width=0.7\textwidth]{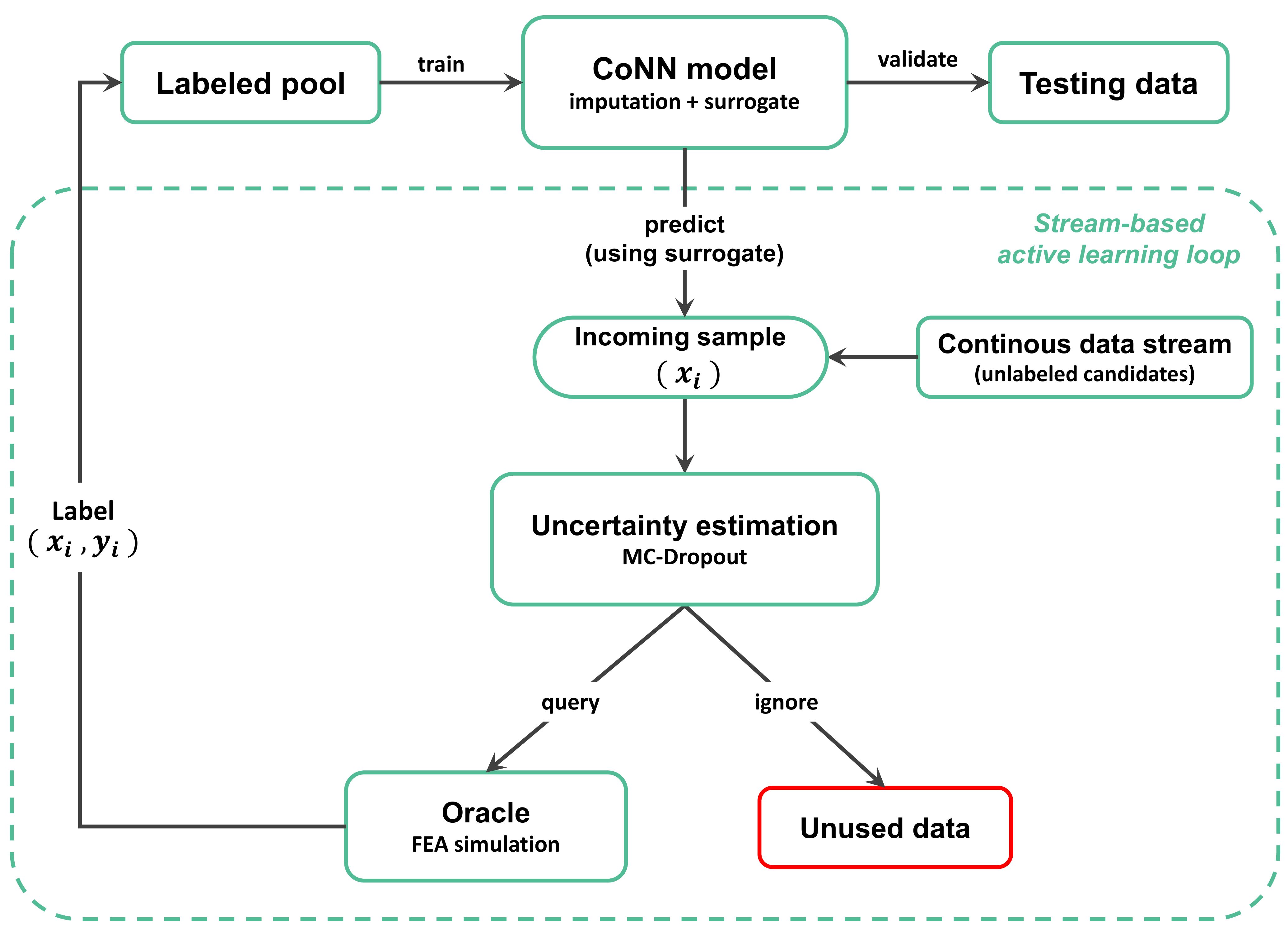}
    \caption{Stream-based active learning loop integrated with CoNN-DAE. Unlabeled samples stream in continuously, and the model estimates predictive uncertainty via MC-Dropout. High-uncertainty samples are queried from the oracle (FEA simulation) and added to the labeled pool, while low-uncertainty samples are discarded.}
    \label{fig:al_loop}
\end{figure}

To reduce the labeling cost associated with simulation-based data generation, this study integrates CoNN-DAE with a stream-based active learning framework, as illustrated in Fig.~\ref{fig:al_loop}. In the stream-based setting, unlabeled samples arrive sequentially from a continuous data source. For each incoming sample $\mathbf{x}_i$, the model must decide in real time whether to query the oracle (finite element simulation) for its label or to discard the sample. This is in contrast to pool-based active learning, where the learner selects from a fixed pool of candidates~\cite{Settles2009_3BU3Z6M2}. The stream-based formulation is a natural fit for simulation-based engineering design, where new candidate designs can be generated on demand and evaluated individually.

The active learning loop proceeds as follows. The CoNN-DAE is first trained on a small initial labeled pool. As new unlabeled samples stream in, the trained model generates predictions using the surrogate component. Predictive uncertainty is then estimated via MC-Dropout~\cite{Gal2016_V3F5AA7Y}, in which $T$ stochastic forward passes are performed with dropout layers active at inference time. The predictive variance for each sample is computed as
\begin{equation}
    \sigma^2(\mathbf{x}_i) = \frac{1}{T} \sum_{t=1}^{T} \left(\hat{y}_{i,t} - \bar{\hat{y}}_i\right)^2
    \label{eq:mc_dropout}
\end{equation}
where $\hat{y}_{i,t}$ is the prediction from the $t$-th forward pass and $\bar{\hat{y}}_i = \frac{1}{T} \sum_{t=1}^{T} \hat{y}_{i,t}$ is the mean prediction. Samples with high predictive variance are considered informative and queried for labeling through the oracle (FEA simulation), while low-uncertainty samples are discarded. The newly labeled samples are added to the training pool, and the model is fine-tuned on the expanded dataset. This process repeats over multiple rounds, progressively building a labeled dataset that is concentrated on the most informative regions of the design space.

By focusing labeling effort on high-uncertainty samples, active learning avoids the redundancy inherent in random sampling, where many acquired samples may fall in well-understood regions of the design space. The specific hyperparameters governing the active learning loop, including the initial pool size, stream batch size, acquisition batch size, uncertainty-to-random sampling ratio, and number of rounds, are detailed in Section~\ref{sec:phases}.

%% ============================================================
%%  SECTION — EXPERIMENT SETUP
%% ============================================================
\section{Experiment Setup}
\label{sec:experiment}

This section describes the dataset, problem formulation, model architecture, experimental protocol, and evaluation metrics used throughout the study. All experiments share the same model, preprocessing, and evaluation pipeline to ensure fair comparison across training strategies.

\subsection{Dataset}
\label{sec:dataset}

\begin{figure}[htbp]
    \centering
    \includegraphics[width=0.8\textwidth]{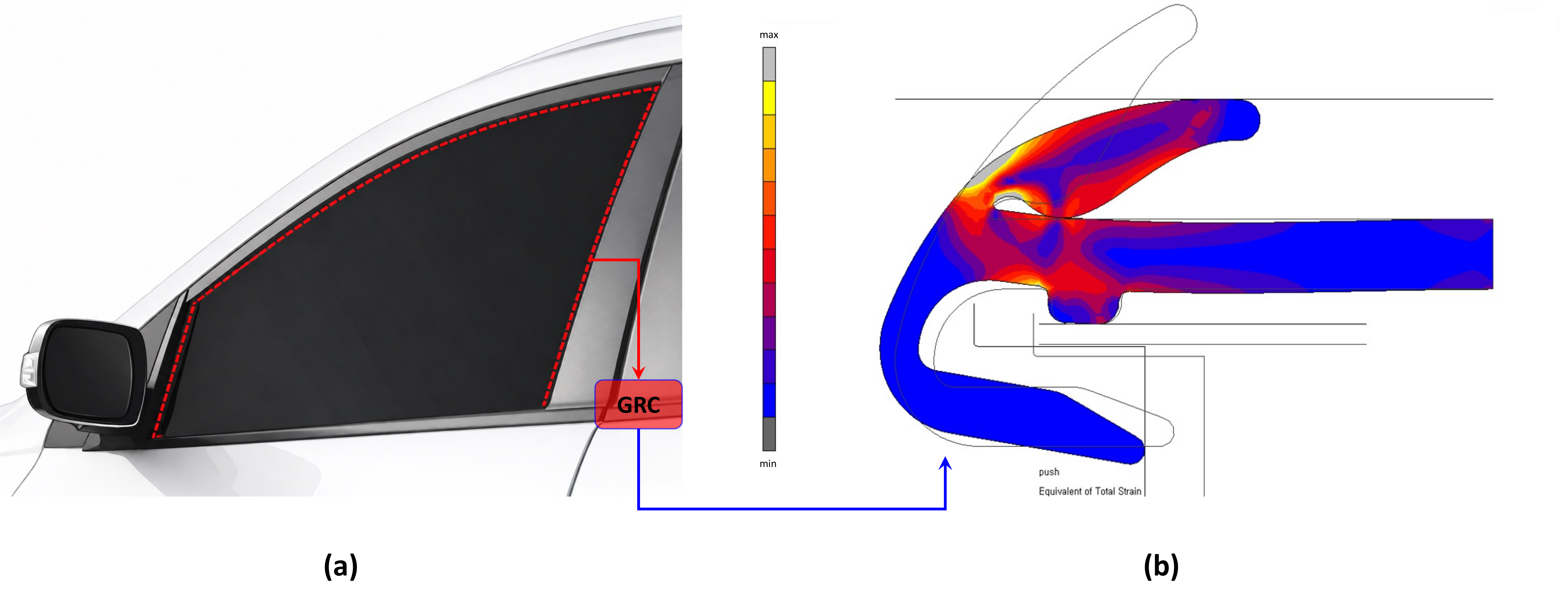}
    \caption{(a) Illustration of the GRC sealing profile location on the vehicle window frame. (b) The GRC profile during finite element analysis in MSC Marc.}
    \label{fig:grc_dataset}
\end{figure}

The Glass Run Channel (GRC) is a rubber sealing component installed along the vehicle window frame to guide and seal the glass panel, as illustrated in Fig.~\ref{fig:grc_dataset}(a). The GRC dataset used in this study was originally introduced and described in Nugraha et al.~\cite{nugraha_cooperative_2025}, developed in collaboration with a Korean automotive company. The original dataset comprises 885,735 unique design records, each parameterized in CATIA~\cite{Mitrache2011_Z5VDMMMF} and evaluated through MSC Marc finite element analysis~\cite{Javanbakht2017_YINUMPL8} (Fig.~\ref{fig:grc_dataset}(b)). For this study, an additional 20,480 samples were generated under the same simulation protocol and appended to the original dataset, bringing the combined total to 906,215 unique records. The full combined dataset is publicly available under the Creative Commons Attribution-NonCommercial-ShareAlike 4.0 International (CC BY-NC-SA 4.0) license at \url{https://doi.org/10.5281/zenodo.23096923}. In the shared version of the dataset, the independent variables are provided in standardized form because of a request that the original design values not be disclosed. The sealing force output is kept in its original physical unit (N/dm).

Each record consists of 12 design variables organized into three functional groups, together with one output variable, as summarized in Table~\ref{tab:grc_variables}. The \textit{Lip} group describes the sealing-lip geometry through four parameters, which are the seal-related gap parameter, the radius of curvature at the lip tip, the radius of curvature at the glass-flat region of the lip profile, and the radius of curvature at the upper-flat region of the lip profile. The \textit{Notch} group characterizes the notch-section geometry through the local lip thickness in the notch region, the notch radius, and two angular parameters, namely the column-notch angle and the notch-lip angle, which govern the relative orientation of the structural features in the sealing cross-section. The \textit{Design} group comprises the remaining structural dimensions, namely column thickness, dam thickness, the position of the inner outer lip, and the radius of curvature of the inner outer lip. The output variable is the sealing force (N/dm), representing the sealing effectiveness of the GRC design under the prescribed finite element analysis conditions.

\begin{table}[htbp]
    \centering
    \small
    \caption{Groups and units of the GRC design variables and output. Units refer to the original measurements; the shared dataset provides the design variables in standardized form.}
    \label{tab:grc_variables}
    \renewcommand{\arraystretch}{1.0}
    \begin{tabular}{llc}
    \toprule
    \textbf{Group} & \textbf{Variable} & \textbf{Unit} \\
    \midrule
    \multirow{4}{*}{Lip Related Data}
        & Seal gap                      & mm           \\
        & Lip tip radius                & mm           \\
        & Lip radius at glass flat      & mm           \\
        & Lip radius at top flat        & mm           \\
    \midrule
    \multirow{4}{*}{Notch Related Data}
        & Notch lip thickness           & mm           \\
        & Notch radius                  & mm           \\
        & Column-notch angle            & $^\circ$     \\
        & Notch-lip angle               & $^\circ$     \\
    \midrule
    \multirow{4}{*}{Design Related Data}
        & Column thickness              & mm           \\
        & Dam thickness                 & mm           \\
        & Inner outer lip position      & mm           \\
        & Inner outer lip radius        & mm           \\
    \midrule
    Output & Sealing force              & N/dm         \\
    \bottomrule
    \end{tabular}
\end{table}

The dataset was partitioned into a training pool of 886,215 unique samples and a held-out test set of 20,000 unique samples. No design combination appears in both the training pool and the test set. The test set size was determined in consultation with domain experts, who confirmed that 20,000 samples provide sufficient coverage of the GRC design space for reliable performance evaluation. The test set was fixed across all experiments to ensure consistent evaluation. Our experiments started from the original recorded values. Before training, the input variables were standardized so that features measured on different scales contributed comparably during gradient-based optimization, and the standardization statistics were computed only from the training pool. The output variable was standardized in the same way, so all error metrics in this study are reported on the standardized output scale. The publicly shared dataset provides the input variables in this standardized form, rounded to three decimals, so the shared data match the inputs used in our experiments. The large scale of this dataset is central to our study. It enables reliable upper-bound benchmarking, reflects the realistic cost of simulation-based labeling in engineering design, and provides broad coverage of the design space for evaluating active learning under varying degrees of missing-variable difficulty.

\subsection{Problem Formulation}
\label{sec:formulation}

This study addresses the partial inverse design problem, in which only a subset of design variables is specified and the remaining variables must be inferred to satisfy a target performance criterion. In practice, certain design parameters may be predetermined by engineering constraints, manufacturing requirements, or regulatory standards, while others remain free for optimization. The task is to reconstruct the unspecified variables such that the completed design achieves the desired sealing force.

To encode this setting, a binary mask matrix (MM) is used to indicate which variables are known (1) and which must be inferred (0), following the formulation in~\cite{nugraha_cooperative_2025}. For each sample, the number of masked variables is drawn randomly between 1 and a specified maximum, and the positions of the masked variables are also selected at random. The parameter MM defines this maximum, ranging from MM$=$1 (at most one variable unknown per sample) to MM$=$6 (up to six variables unknown out of twelve). Because both the count and identity of the missing variables vary randomly across samples, the model must learn to reconstruct arbitrary subsets of design variables rather than a fixed missing pattern. Increasing MM progressively raises the difficulty of the inference task, as the model encounters a wider range of incomplete input specifications. Throughout this paper, the bold symbol $\mathbf{MM}$ denotes the binary mask matrix itself, while the MM level (MM$=$1 to MM$=$6) denotes the maximum number of missing variables per sample. This formulation enables systematic evaluation of model robustness under varying degrees of design incompleteness.

\subsection{Evaluation Configuration}
\label{sec:phases}

The experiments are organized into four stages, each designed to address a specific question about data efficiency and model performance under partial inverse design conditions. All experiments use the CoNN-DAE architecture as described in Section~\ref{sec:methodology}.

\subsubsection{Upper-Bound Benchmark}
The first stage establishes the performance ceiling by training CoNN-DAE on two large labeled subsets of 100,000 samples and the full training pool of 886,215 samples. Each model is trained for 100 epochs. These upper-bound models serve as reference points against which data-efficient strategies are compared. Performance is evaluated separately for each MM level (MM$=$1 to MM$=$6) to characterize the relationship between training set size and prediction quality under varying degrees of incomplete input specification.

\subsubsection{Random Sampling Baseline}
The second stage constructs learning curves using random sampling as a passive data acquisition baseline. Models are trained for 100 epochs on randomly selected subsets of increasing size, from 10,000 to 100,000 samples in steps of 10,000, drawn from the training pool. Each training size is evaluated separately for each MM level (MM$=$1 to MM$=$6). This stage quantifies how model performance scales with dataset size under uninformed (random) selection, establishing the baseline against which active learning is compared.

\subsubsection{Stream-Based Active Learning}
The third stage evaluates stream-based active learning as an informed data acquisition strategy, conducted separately for each MM level (MM$=$1 to MM$=$6). The active learning loop proceeds as follows. An initial seed pool of 3,000 randomly selected labeled samples is used to train the CoNN-DAE. In each subsequent round, a stream batch of 3,000 candidate samples arrives from the unlabeled stream. The 750 candidates with the highest predictive uncertainty (estimated via MC-Dropout with $T=20$ forward passes) and 250 randomly chosen candidates are queried from the oracle, while the remaining 2,000 candidates are discarded and do not return to the stream. Each round therefore adds 1,000 labeled samples to the training pool. After each acquisition, the model is fine-tuned from its previous weights for 10 additional epochs. This incremental fine-tuning schedule differs from the single 100-epoch training used in the upper-bound and random sampling stages, so the reported comparisons represent the performance of the complete acquisition-and-training protocols at matched labeling budgets. This process is repeated for 100 rounds, yielding a final labeled pool of approximately 103,000 samples. Because the model is evaluated in each round before that round's acquisition, the model evaluated at round $r$ has been trained on $3{,}000 + 1{,}000\,(r-1)$ labeled samples. Key checkpoints are examined at round 18 (20,000 labels), round 74 (76,000 labels), and round 100 (102,000 labels).

\subsubsection{Label Efficiency Analysis}
The fourth stage directly compares the three strategies. Active learning, random sampling, and upper-bound training are compared under matched labeling budgets. Two comparison points are examined, (a) a fixed budget of 20,000 labeled samples, and (b) the most efficient active learning checkpoint at approximately 76,000 labels. Additionally, label efficiency is quantified by measuring the number of labeled samples each strategy requires to reach specific $R^2$ thresholds under the most challenging scenario (MM$=$6).

\subsection{Evaluation Metrics}
\label{sec:metrics}

Model performance is assessed using five complementary metrics, summarized in Table~\ref{tab:metrics}. All metrics are computed on the held-out test set of 20,000 samples and reported separately for each MM level (MM$=$1 through MM$=$6) to characterize performance across different degrees of incomplete input specification.

\begin{table}[htbp]
    \centering
    \small
    \caption{Evaluation metrics and their definitions.}
    \label{tab:metrics}
    \renewcommand{\arraystretch}{1.4}
    \begin{tabular}{>{\raggedright\arraybackslash}p{6cm} c}
    \toprule
    \textbf{Metric} & \textbf{Definition} \\
    \midrule
    Coefficient of Determination ($R^2$)
    & $1 - \dfrac{\sum_{i=1}^{m} (y_i - \hat{y}_i)^2}{\sum_{i=1}^{m} (y_i - \bar{y})^2}$ \\[6pt]
    Mean Squared Error (MSE)
    & $\dfrac{1}{m} \sum_{i=1}^{m} (y_i - \hat{y}_i)^2$ \\[6pt]
    Mean Absolute Error (MAE)
    & $\dfrac{1}{m} \sum_{i=1}^{m} \left| y_i - \hat{y}_i \right|$ \\[6pt]
    Root Mean Squared Error (RMSE)
    & $\sqrt{\dfrac{1}{m} \sum_{i=1}^{m} (y_i - \hat{y}_i)^2}$ \\[6pt]
    Symmetric Mean Absolute Percentage Error (sMAPE)
    & $\dfrac{2}{m} \displaystyle\sum_{i=1}^{m} \dfrac{\left| y_i - \hat{y}_i \right|}{\left| y_i \right| + \left| \hat{y}_i \right|}$ \\
    \bottomrule
    \end{tabular}
\end{table}

%% ============================================================
%%  SECTION — RESULTS AND DISCUSSION
%% ============================================================

\section{Results and Discussion}
\label{sec:results}

This section presents the experimental results across the four phases described in Section~\ref{sec:experiment}. The overarching finding is that CoNN-DAE is naturally suited to partial inverse design under incomplete input specification, and that stream-based active learning amplifies its data efficiency, enabling near upper-bound performance with a fraction of the labeling cost that would otherwise require extensive CAE simulation. The results are organized to support three contributions, namely (A) data efficiency of active learning, (B) near upper-bound performance with limited labels, and (C) robustness across partial inverse design difficulty levels. The fourth contribution stated in Section~\ref{sec:introduction}, the public release of the GRC dataset, is addressed in Section~\ref{sec:dataset} and is therefore not evaluated here.

\subsection{Upper-Bound Benchmark Performance}
\label{sec:results_ub}

The upper-bound experiment establishes the performance ceiling for CoNN-DAE on the GRC dataset by training on 100,000 and 886,215 labeled samples. In the original CoNN study~\cite{nugraha_cooperative_2025}, training on approximately 10\% of the available data already yielded strong predictive performance, demonstrating the model's capacity to learn effectively from limited supervision. However, even a 10\% subset of the GRC dataset corresponds to roughly 90,000 labeled designs, each requiring CATIA modeling and MSC Marc finite element evaluation, a process that demands substantial computational resources. This cost motivates the investigation of active learning as a strategy for targeted, efficient data acquisition. The reliability of CoNN-DAE under limited data is precisely the property that makes this integration naturally promising. If the model already performs well with small datasets, then intelligently selecting which samples to label should amplify this advantage further.

\begin{figure}[htbp]
    \centering
    \includegraphics[width=0.6\textwidth]{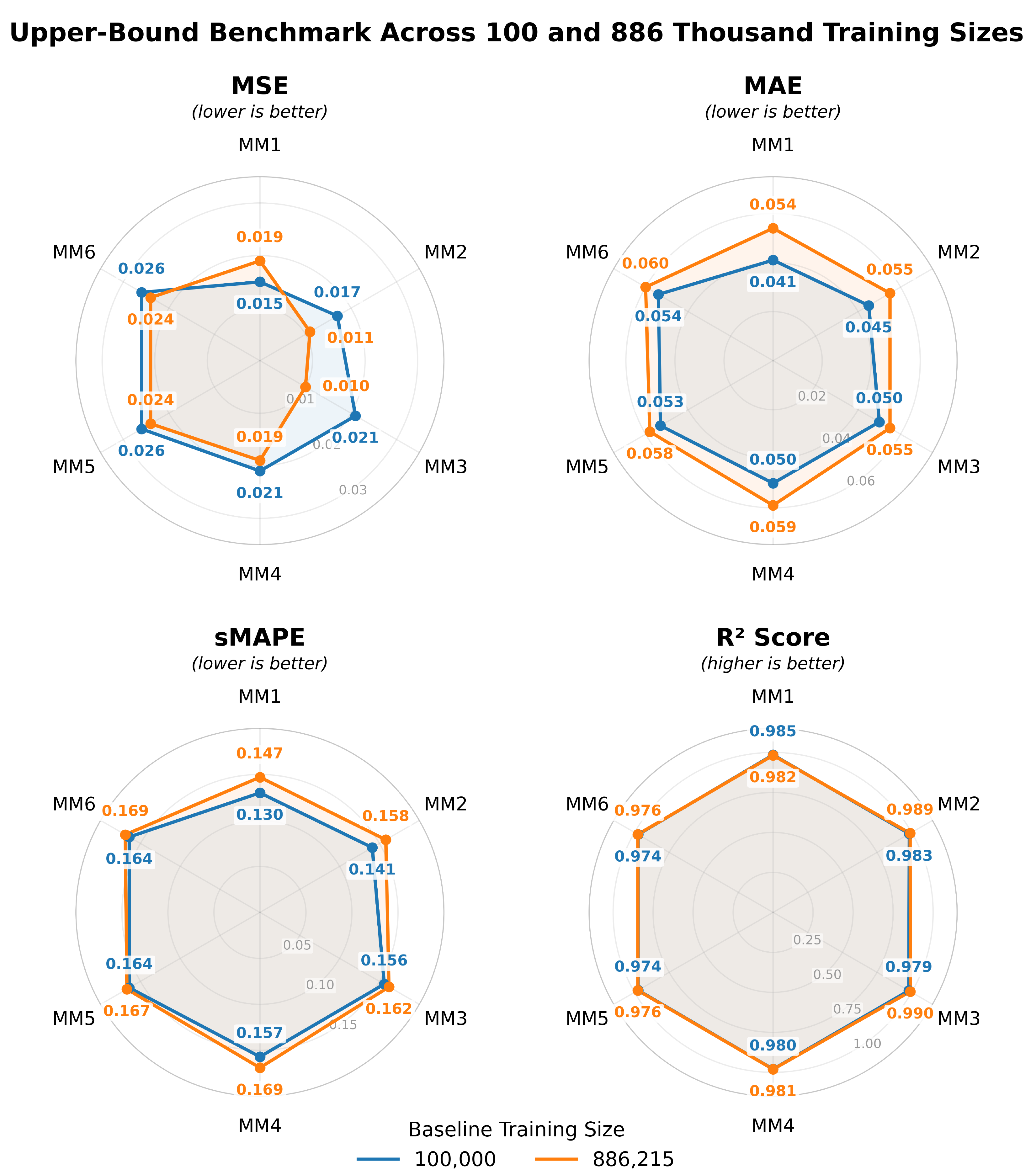}
    \caption{Radar chart comparison of upper-bound performance metrics (MSE, MAE, sMAPE, $R^2$) between 100,000 and 886,215 training sizes across MM levels. Performance is largely saturated at 100,000 samples.}
    \label{fig:ub_radar}
\end{figure}

A side-by-side comparison of MSE, MAE, sMAPE, and $R^2$ across the two training sizes for each MM level, as shown in the radar charts of Fig.~\ref{fig:ub_radar}, reveals that the two training sizes trace very similar polygons in most panels, with visible gaps only in the MSE panel at MM$=$2 and MM$=$3. In the $R^2$ panel, both training sizes consistently achieve values between 0.97 and 0.99, with differences of about one percentage point or less at every MM level. The MSE panel shows that the 886,215 model achieves lower error at most MM levels, most visibly at MM$=$2 and MM$=$3, while at MM$=$1 the 100,000 model performs slightly better. In contrast, the MAE and sMAPE panels show marginally lower values for the 100,000 model across all MM levels. This near-equivalence across training sizes indicates that model performance is already saturated at 100,000 labeled samples, nearly an order of magnitude less than the full dataset. This saturation is a direct consequence of the cooperative learning mechanism in CoNN-DAE, which enables efficient utilization of available supervision. The observed differences are sufficiently small to be practically negligible in downstream engineering use. In other words, both models provide effectively comparable predictive quality despite the large difference in labeling budget. From a practical standpoint, this result highlights the suitability of CoNN-DAE for data-efficient partial inverse design, confirming contribution (B), namely that the model approaches its performance ceiling well before the full labeling budget is exhausted.

\begin{figure}[htbp]
    \centering
    \includegraphics[width=0.9\textwidth]{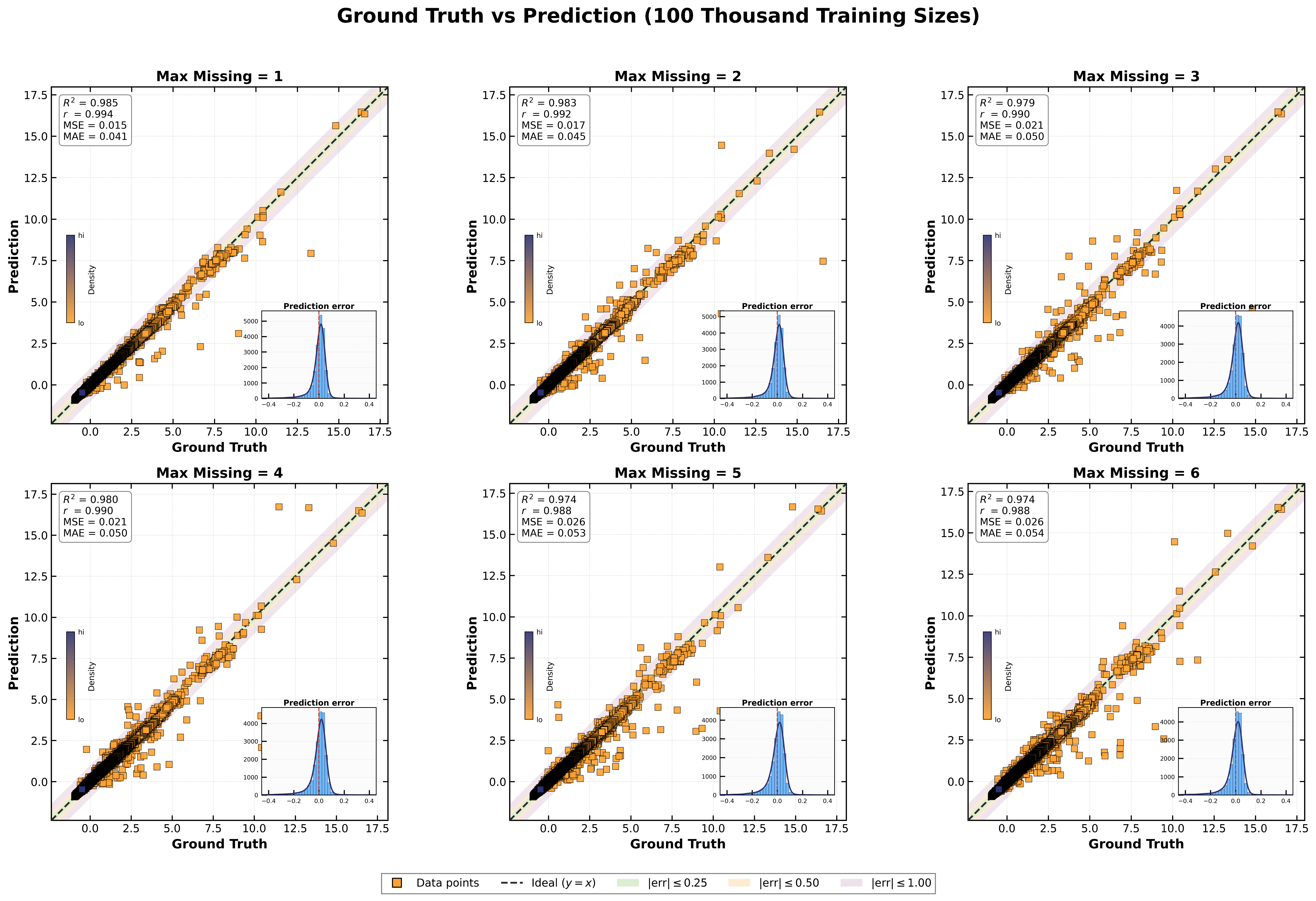}
    \caption{Ground truth vs.\ prediction scatter plots for CoNN-DAE trained on 100,000 samples (MM$=$1 to MM$=$6). Points align closely with the identity line across all MM levels.}
    \label{fig:ub_scatter_100K}
\end{figure}

\begin{figure}[htbp]
    \centering
    \includegraphics[width=0.9\textwidth]{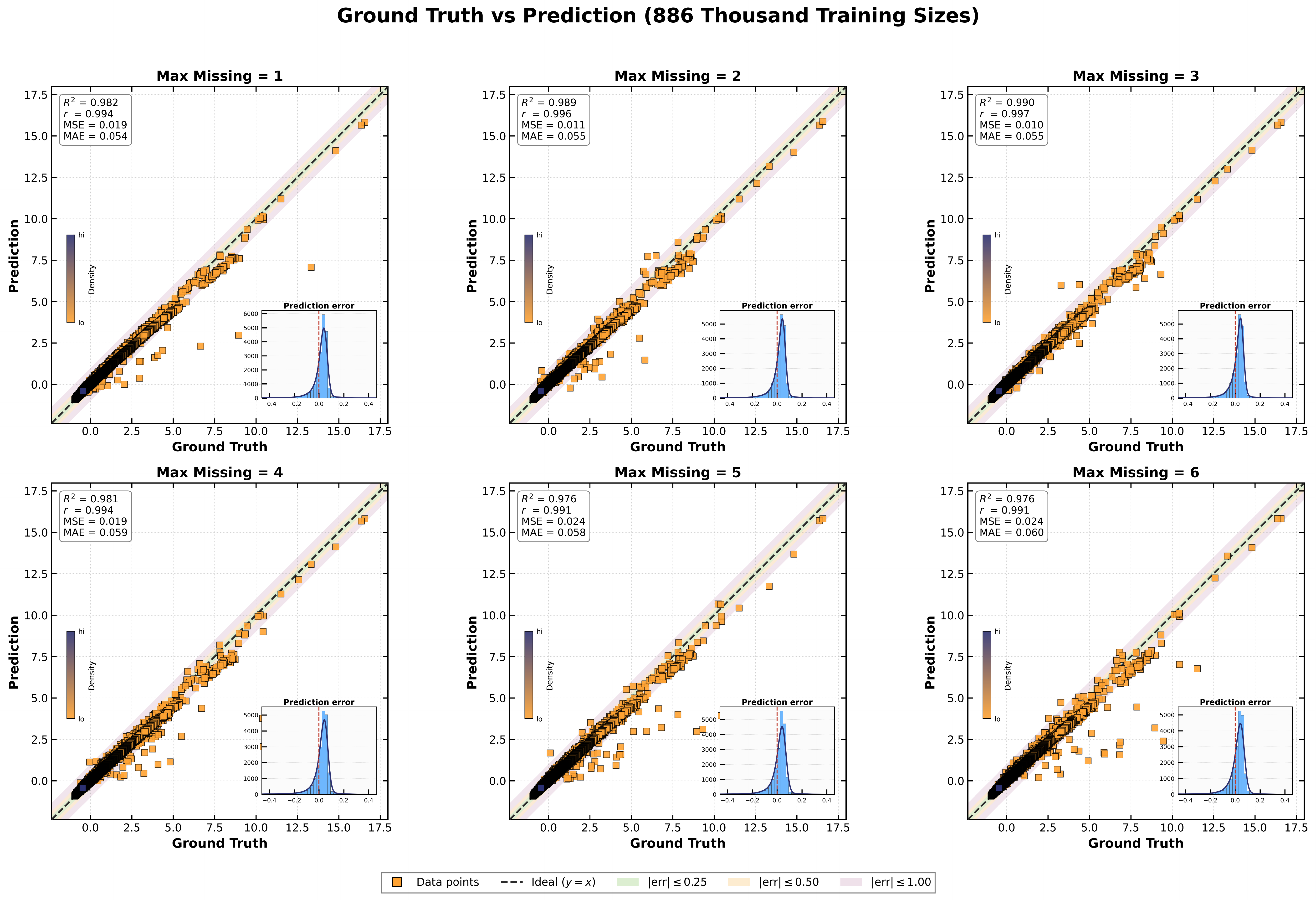}
    \caption{Ground truth vs.\ prediction scatter plots for CoNN-DAE trained on 886,215 samples (MM$=$1 to MM$=$6). The pattern closely matches the 100,000 model, confirming performance saturation.}
    \label{fig:ub_scatter_886K}
\end{figure}

Figures~\ref{fig:ub_scatter_100K} and~\ref{fig:ub_scatter_886K} show the ground-truth versus predicted scatter plots for the 100,000 and 886,215 models, respectively, with one subplot for each MM level. In both figures, the predictions align closely with the identity line, indicating accurate estimation across the full output range. As MM increases from 1 to 6, the scatter shows a modest increase in dispersion and a greater number of off-diagonal points, reflecting the increasing difficulty of reconstructing a larger number of missing design variables from more incomplete inputs. However, this degradation remains gradual rather than severe. Even at MM$=$6, the predictions remain concentrated near the diagonal and maintain $R^2$ values above 0.97. Moreover, the overall scatter patterns for the 100,000 and 886,215 models are visually very similar, providing further evidence that performance is already saturated at 100,000 labeled samples. This stable behavior is consistent with the findings of the original CoNN study~\cite{nugraha_cooperative_2025}, where the cooperative DAE framework remained robust as the number of undetermined variables increased and continued to outperform standalone imputation models under more challenging conditions. These results support contribution (C), showing that CoNN-DAE remains robust across varying levels of partial inverse design difficulty, while the mild increase in dispersion at higher MM levels is likely associated with the greater intrinsic difficulty of the task.

\subsection{Random Sampling Baseline}
\label{sec:results_random}

\begin{figure}[htbp]
    \centering
    \includegraphics[width=1.0\textwidth]{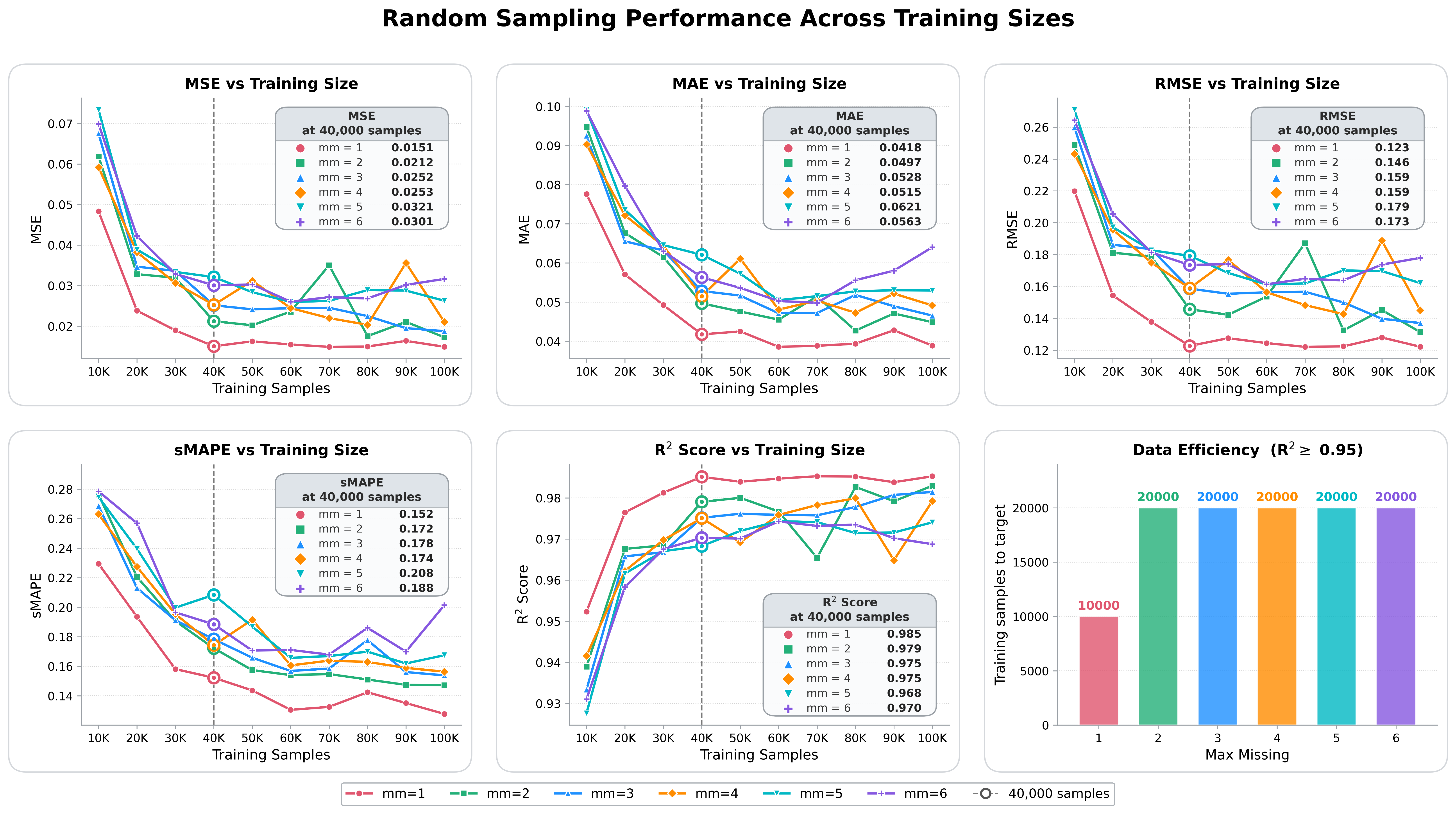}
    \caption{Learning curves for random sampling baseline across all MM levels. Performance metrics (MSE, MAE, RMSE, sMAPE, $R^2$) are shown as a function of training size (10,000--100,000). Insets report each metric at the 40,000-sample checkpoint (dashed line). The bottom-right panel shows the number of training samples required to reach $R^2 \geq 0.95$ at each MM level. A performance plateau is reached around 40,000 samples.}
    \label{fig:random_curves}
\end{figure}

The learning curves for the random sampling baseline across all six MM levels are shown in Fig.~\ref{fig:random_curves}, covering MSE, MAE, RMSE, sMAPE, and $R^2$ as a function of training size. Each metric panel includes an inset bar chart reporting the corresponding value at the 40,000-sample checkpoint, marked by the dashed vertical line, and the bottom-right panel summarizes data efficiency as the number of training samples required at each MM level to reach $R^2 \geq 0.95$. The overall trend is highly consistent across MM levels. All error metrics drop sharply from 10,000 to 30,000 samples, showing that the model benefits most from the initial increase in training data. After this early improvement phase, the gains become progressively smaller, and the curves reach an evident plateau at around 40,000 samples. This behavior indicates diminishing returns from additional randomly selected labels once the major structure of the design space has already been learned. The inset bar charts quantify this plateau. At the 40,000-sample checkpoint, the model attains $R^2$ values between 0.968 (MM$=$5) and 0.985 (MM$=$1), with the most difficult setting, MM$=$6, at 0.970. These values are close to the upper-bound benchmark although less than half of the 100,000-sample budget is used, confirming that random sampling reaches a practical saturation region relatively early and that additional randomly selected data contribute increasingly limited performance gains after the initial stage.

This plateau behavior holds consistently across all MM levels, suggesting that the relationship between dataset size and model performance under random sampling is relatively stable with respect to the degree of incomplete input specification. Although several non-monotonic fluctuations appear at specific size and MM combinations, the differences are very small in magnitude and have little practical effect on the overall interpretation of the results. For example, the MSE curve for MM$=$2 increases slightly around 70,000 samples before improving again at 80,000, and MM$=$4 shows a temporary error increase around 90,000 before recovering at 100,000. Such minor variations are reasonable when comparing independently sampled random subsets, especially after performance has already approached the practical saturation region near the upper-bound benchmark. Taken together, the overall trend remains clear and consistent across MM levels.

The plateau around 40,000 samples, together with the observed non-monotonic behavior, provides a clear motivation for active learning. The data efficiency panel of Fig.~\ref{fig:random_curves} shows that random sampling requires 10,000 samples to reach $R^2 \geq 0.95$ at MM$=$1 and 20,000 samples at every other MM level (MM$=$2 to MM$=$6). These values establish the reference labeling cost against which the active learning strategy is evaluated in Section~\ref{sec:results_al}. Since performance enters a practical saturation region by around 40,000 randomly selected samples, an informed acquisition strategy that prioritizes more informative samples may achieve comparable performance with fewer labels, or improved performance under the same labeling budget. In the GRC setting, each labeled sample requires CATIA-based design generation and MSC Marc simulation. Therefore, the diminishing returns beyond 40,000 samples indicate that additional labeling cost under random sampling yields only limited benefit. These diminishing returns motivate contribution (A), indicating that a targeted acquisition strategy may achieve comparable performance at a substantially lower labeling cost, which is examined in the following subsections.

\subsection{Stream-Based Active Learning}
\label{sec:results_al}

\begin{figure}[htbp]
    \centering
    \includegraphics[width=1.0\textwidth]{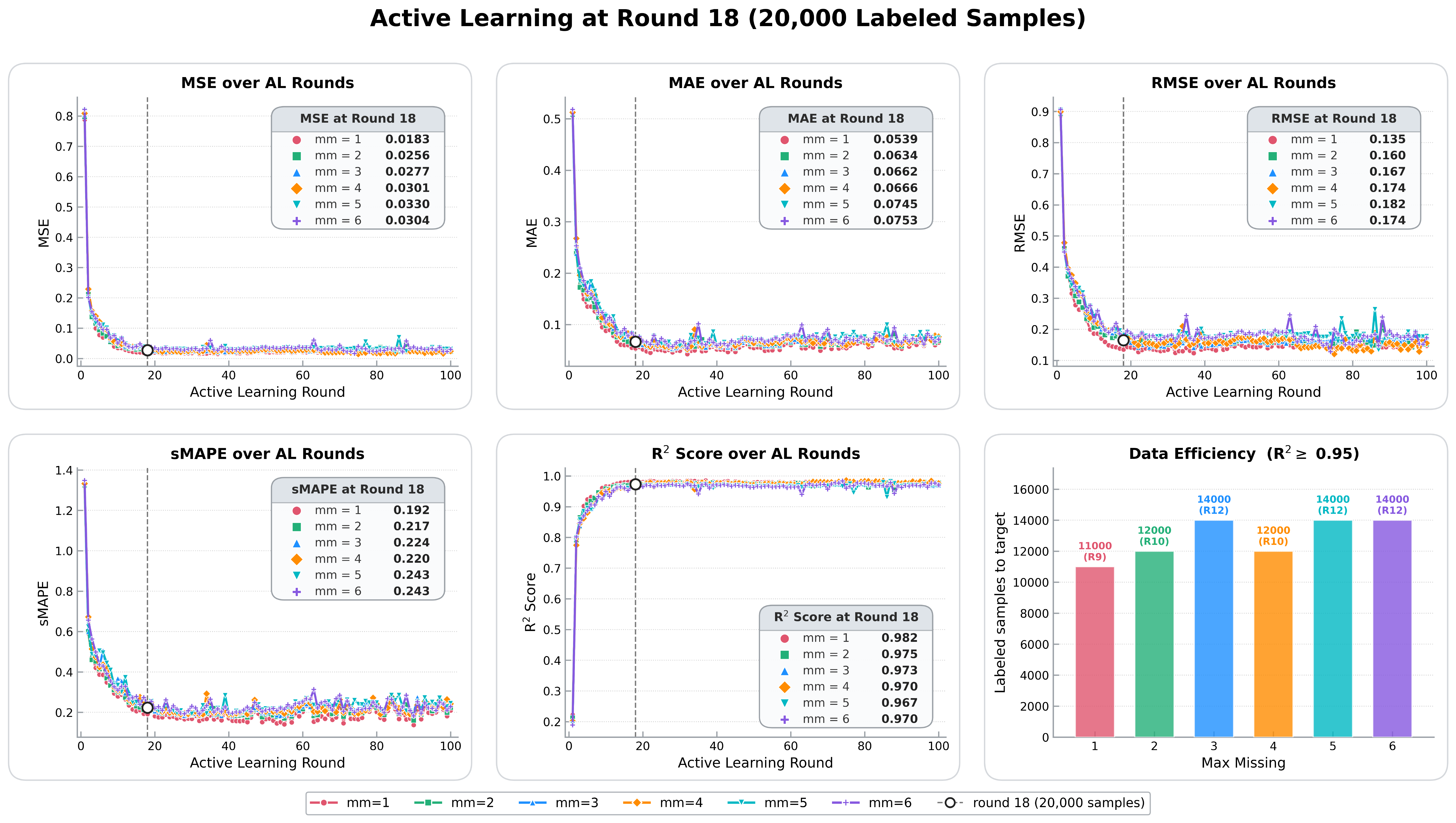}
    \caption{Active learning convergence curves (MSE, MAE, RMSE, sMAPE, $R^2$) over 100 rounds, with insets reporting each metric at round 18 (20,000 labeled samples), marked by the dashed line. The bottom-right panel shows the number of labeled samples, and the corresponding round, required to reach $R^2 \geq 0.95$ at each MM level. Performance is consistent across all MM levels at this early checkpoint.}
    \label{fig:al_curves}
\end{figure}

The most notable feature of the CoNN-AL active learning process, as illustrated in Fig.~\ref{fig:al_curves}, is the steep improvement in all metrics during the first 10--15 rounds, after which the curves level off into a stable plateau. The figure reports MSE, MAE, RMSE, sMAPE, and $R^2$ over 100 acquisition rounds. Insets report each metric at round 18, marked by the dashed vertical line, and the bottom-right panel summarizes data efficiency as the number of labeled samples, and the corresponding round, at which each MM level reaches $R^2 \geq 0.95$. This rapid early convergence reflects the core mechanism of active learning. By prioritizing high-uncertainty samples, the model acquires the most informative examples first. Steep early improvement also occurs under random sampling, so the strength of this mechanism is confirmed separately by the matched-budget comparison in Section~\ref{sec:results_efficiency}, where active learning reaches higher accuracy than random sampling at the same 20,000-label budget. By round 18, with only 20,000 labeled samples, $R^2$ values already range from 0.967 to 0.982 across all MM levels, while MSE ranges from 0.0183 to 0.0330 and sMAPE remains below 0.25 at all MM levels. The consistency of these values across increasing MM demonstrates that the uncertainty-guided acquisition strategy selects samples that are informative across the full spectrum of design incompleteness. Even under the most challenging scenario (MM$=$5 and MM$=$6), the model maintains $R^2$ of at least 0.967, supporting contribution (C), namely that robustness across partial inverse design difficulty levels is preserved even at low labeling budgets.

The data efficiency panel further quantifies the labeling cost of active learning relative to the random sampling baseline established in Section~\ref{sec:results_random}. Active learning reaches $R^2 \geq 0.95$ with 11,000 labeled samples (round 9) at MM$=$1, 12,000 labels (round 10) at MM$=$2 and MM$=$4, and 14,000 labels (round 12) at MM$=$3, MM$=$5, and MM$=$6. In contrast, random sampling requires 20,000 samples to reach the same threshold at every MM level except MM$=$1 (Fig.~\ref{fig:random_curves}). Active learning therefore reduces the labeling cost to this threshold by 30--40\% at the five more difficult MM levels, while the two strategies remain comparable at MM$=$1 (10,000 versus 11,000). This pattern is consistent with the difficulty of the inference task. When only one variable is missing, most candidate samples are informative and uncertainty-guided selection provides limited additional value, whereas at higher MM levels informative samples become scarcer and the benefit of uncertainty-based acquisition increases accordingly. This threshold result supports contribution (A) from the early rounds of the acquisition process.

\begin{figure}[htbp]
    \centering
    \includegraphics[width=0.9\textwidth]{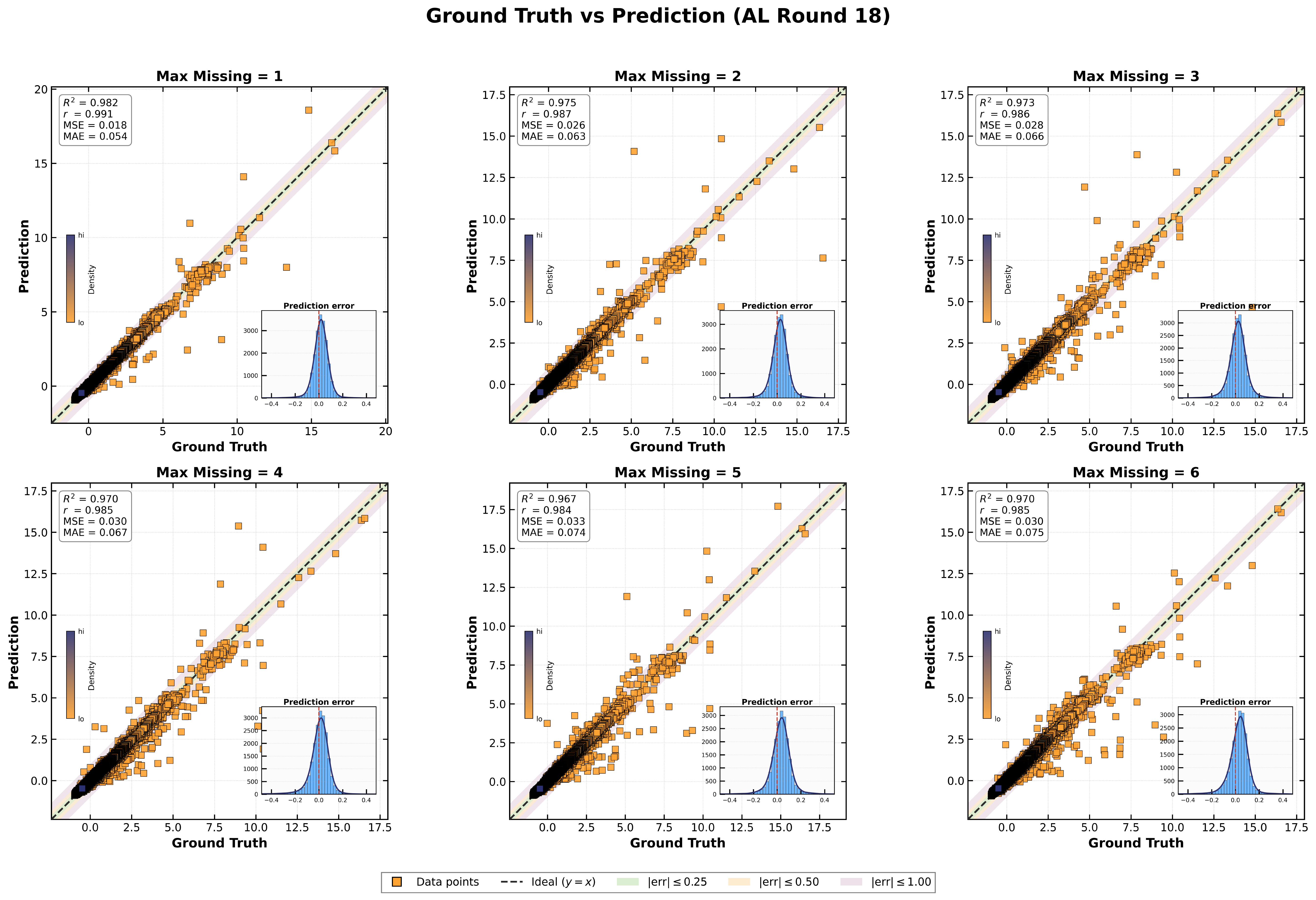}
    \caption{Ground truth vs.\ prediction scatter plots at active learning round 18 (20,000 labels) for MM$=$1 to MM$=$6. Near upper-bound prediction accuracy is achieved with approximately 2.3\% of the full labeling budget.}
    \label{fig:al_scatter_r18}
\end{figure}

Figure~\ref{fig:al_scatter_r18} examines prediction quality at round 18 through ground-truth versus predicted scatter plots for each MM level. Across all six MM levels, the predictions align closely with the identity line, achieving $R^2$ values of 0.982 (MM$=$1), 0.975 (MM$=$2), 0.973 (MM$=$3), 0.970 (MM$=$4), 0.967 (MM$=$5), and 0.970 (MM$=$6). These values approach the upper-bound benchmark (Figs.~\ref{fig:ub_radar}, \ref{fig:ub_scatter_100K}, and~\ref{fig:ub_scatter_886K}) despite using only 20,000 labeled samples, roughly 2.3\% of the full training pool of 886,215. In the original CoNN study~\cite{nugraha_cooperative_2025}, assembling the full GRC dataset of 885,735 designs required approximately three weeks of continuous CATIA/MSC Marc simulation. The active learning result at round 18 demonstrates that comparable predictive performance is achievable with a fraction of that labeling effort when guided by informative sample selection. This supports contribution (B), namely near upper-bound performance with limited labels, and contribution (A), namely the data efficiency advantage of active learning over exhaustive labeling.

\begin{figure}[htbp]
    \centering
    \includegraphics[width=1.0\textwidth]{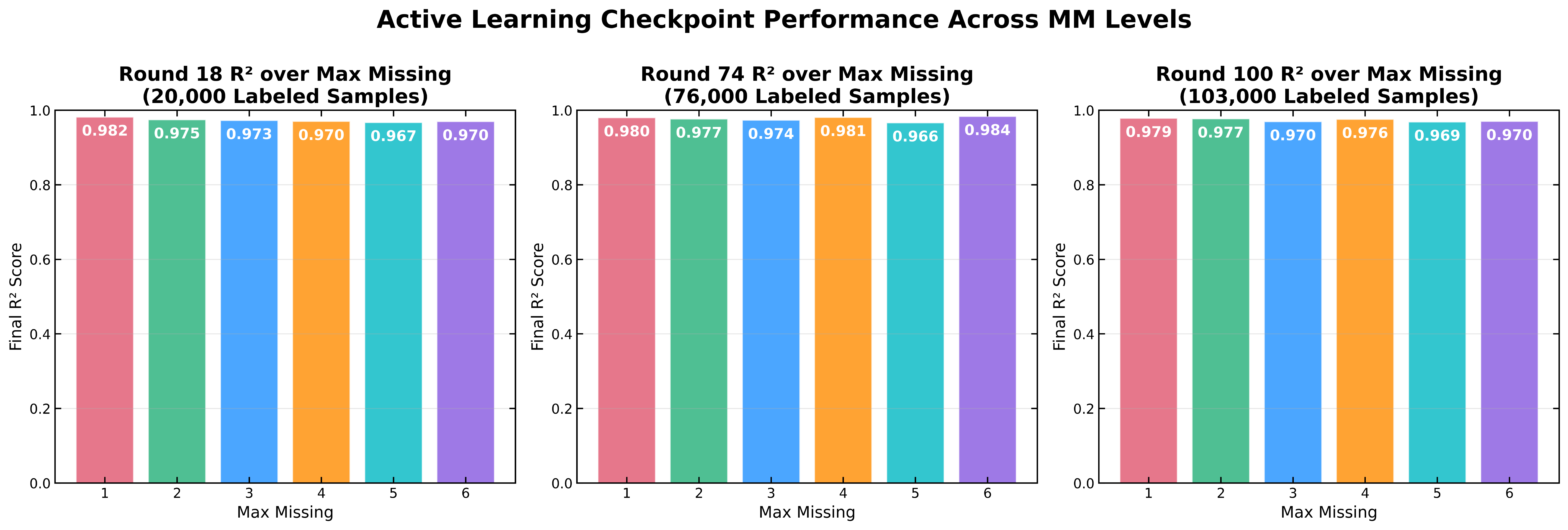}
    \caption{$R^2$ comparison at three active learning checkpoints (round 18, round 74, round 100) across MM levels. Round 74 (76,000 labeled samples) represents the optimal stopping point.}
    \label{fig:al_checkpoints}
\end{figure}

To examine how performance evolves across the active learning process, $R^2$ is compared at three key checkpoints, round 18 (20,000 labeled samples), round 74 (76,000 labeled samples), and round 100 (102,000 labeled samples), as shown in Fig.~\ref{fig:al_checkpoints}. The progression across checkpoints reveals the dynamics of active learning convergence. At round 18, the model already achieves $R^2$ values of 0.967--0.982, indicating that most of the learning occurs in the early rounds when the most informative samples are acquired. By round 74, several MM levels reach their highest observed $R^2$ values, notably MM$=$4 (0.981) and MM$=$6 (0.984), which are higher than the corresponding round 100 results of 0.976 and 0.970. This indicates that acquisition beyond 76,000 samples brings no further improvement and even causes a mild decline, about 1.3 percentage points at MM$=$6, likely because the later stream batches contain increasingly redundant samples. The round 100 checkpoint still maintains $R^2$ values of 0.969--0.979 across all MM levels, so the decline has limited practical impact, yet it strengthens the case for stopping the acquisition early. Taken together, these results suggest that round 74 (76,000 labeled samples) represents an efficient stopping point, where the labeling effort can be reduced by approximately 91\% relative to the full training pool (76,000 vs.\ 886,215), or by 24\% relative to the 100,000-sample upper-bound budget, without sacrificing predictive quality. This finding supports contributions (A) and (B), showing that active learning accelerates convergence through efficient sample selection and identifies a practical stopping point that balances performance with labeling cost.

\subsection{Label Efficiency and Comparative Analysis}
\label{sec:results_efficiency}

To rigorously evaluate the label efficiency advantage of active learning, this section presents direct comparisons between active learning, random sampling, and the upper-bound models under matched labeling budgets.

\begin{figure}[htbp]
    \centering
    \includegraphics[width=1.0\textwidth]{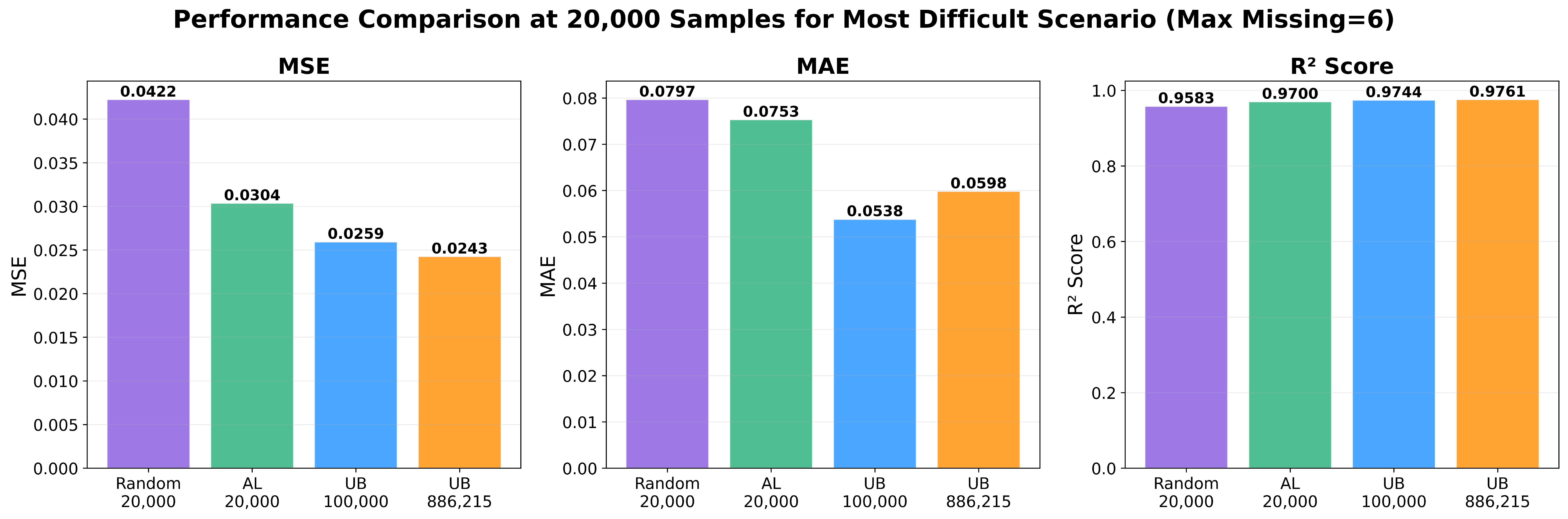}
    \caption{Performance comparison of active learning, random sampling, and the upper-bound models (100,000 and 886,215) at 20,000 labeled samples (MM$=$6). Active learning outperforms random sampling and approaches the upper-bound.}
    \label{fig:matched_20K}
\end{figure}

Under a fixed budget of 20,000 labeled samples and the most challenging scenario (MM$=$6), the three strategies show distinct performance levels, as compared in Fig.~\ref{fig:matched_20K}. Active learning achieves an $R^2$ of 0.970, clearly outperforming random sampling at 0.958 (a difference of 1.2 percentage points) and approaching the upper-bound models at 0.974 (100,000) and 0.976 (886,215). In terms of MSE, active learning achieves 0.030 compared to 0.042 for random sampling, a reduction of approximately 28\%. The MAE comparison shows a similar pattern, with active learning at 0.075 versus random sampling at 0.080. With the same label budget, active learning substantially outperforms random sampling and approaches the upper-bound, demonstrating superior label utilization in partial inverse design. This advantage arises because the uncertainty-guided acquisition strategy concentrates labels on the most informative regions of the design space, whereas random sampling allocates labels uniformly regardless of their informational value. This result directly supports contributions (A) and (B) that active learning is more data-efficient and achieves near upper-bound performance with limited labels.

\begin{figure}[htbp]
    \centering
    \includegraphics[width=1.0\textwidth]{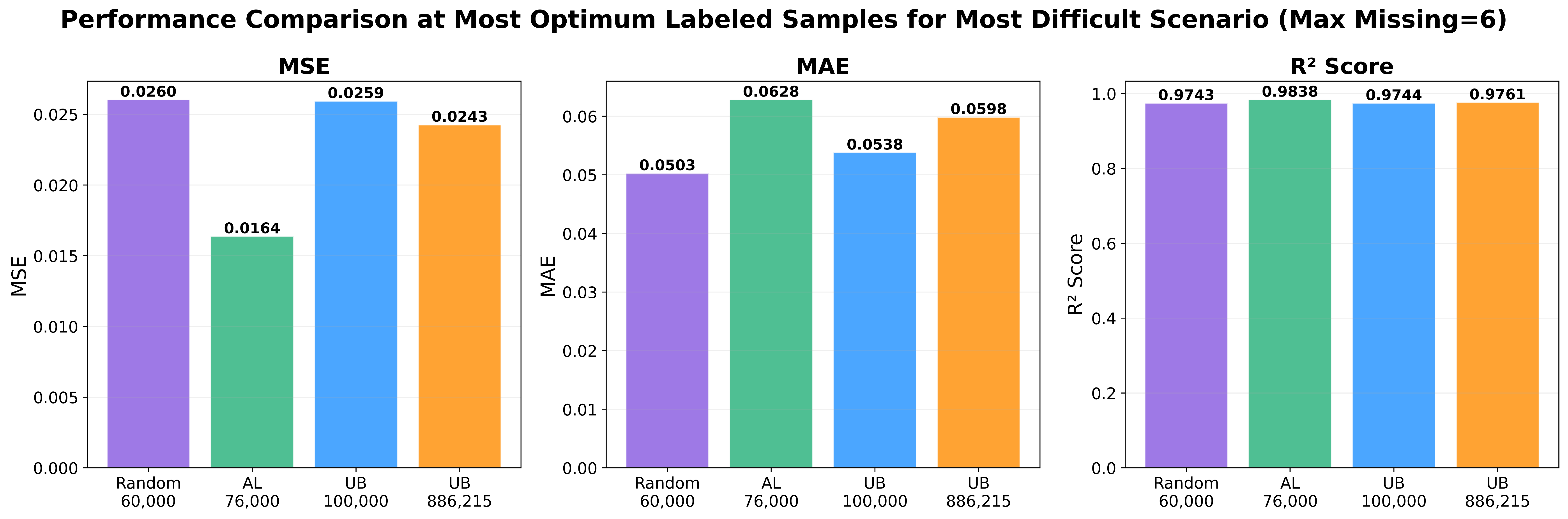}
    \caption{Performance comparison at the most efficient active learning checkpoint (76,000 labels) vs.\ random sampling (60,000) and upper-bound (100,000, 886,215) under MM$=$6. Active learning achieves lower MSE and slightly higher $R^2$ than the upper-bound models, while MAE is higher.}
    \label{fig:matched_optimal}
\end{figure}

Figure~\ref{fig:matched_optimal} extends this comparison to the most efficient active learning checkpoint at approximately 76,000 labels, evaluated against the best random sampling result at 60,000 labels and the upper-bound models. The 60,000-label point is chosen because it yields the highest $R^2$ among all random sampling sizes; the random learning curve fluctuates after its plateau, and larger random subsets up to 100,000 labels do not exceed this result. In terms of $R^2$, active learning at 76,000 achieves 0.9838, which is marginally higher than random sampling at 60,000 (0.9743), UB 100,000 (0.9744), and UB 886,215 (0.9761). While the $R^2$ differences are small in absolute terms, the MSE comparison reveals a more pronounced advantage. Active learning achieves MSE of 0.0164, which is approximately 37\% lower than both random sampling (0.0260) and the 100,000 upper-bound model (0.0259), and approximately 32\% lower than the 886,215 upper-bound model (0.0243). This suggests that uncertainty-guided sample selection is particularly effective at reducing large prediction errors. However, the MAE comparison presents a more nuanced picture. Active learning achieves a MAE of 0.0628, which is higher than random sampling (0.0503), UB 100,000 (0.0538), and UB 886,215 (0.0598). This indicates that while active learning reduces the magnitude of large errors more effectively, it does not necessarily reduce average absolute errors uniformly across all samples. This mixed pattern highlights the importance of considering multiple evaluation metrics, as the choice of metric can influence how the benefit of active learning is interpreted. Overall, active learning trained on 76,000 carefully selected samples achieves comparable or better performance across most metrics relative to models trained on larger but randomly composed datasets, using fewer labeled samples. These results support contributions (A) and (B), showing that targeted sample selection can match or modestly improve upon passive strategies with a reduced labeling budget. Notably, this advantage persists under the most challenging MM$=$6 scenario, consistent with the robustness stated in contribution (C).

Label efficiency is further quantified in Table~\ref{tab:label_efficiency} by reporting the number of labeled samples each strategy requires to reach specific $R^2$ thresholds under the most challenging MM$=$6. At lower thresholds ($R^2 = 0.90$), both strategies perform comparably. However, as the target $R^2$ increases, the efficiency gap widens. Reaching $R^2 = 0.96$ requires 30,000 randomly sampled labels but only 15,000 actively selected labels (2.0$\times$ efficiency), and reaching $R^2 = 0.97$ requires 40,000 random labels versus 20,000 active learning labels (2.0$\times$ efficiency). Notably, the $R^2 = 0.98$ threshold is never reached by random sampling within 100,000 labels, while active learning achieves it at 76,000 labels. This threshold analysis provides a concrete, interpretable measure of the labeling cost advantage conferred by active learning. These results are consistent with the data efficiency panels in Figs.~\ref{fig:random_curves} and~\ref{fig:al_curves}, which report the same $R^2 \geq 0.95$ crossing points (20,000 for random sampling versus 14,000 for active learning at MM$=$6).

\begin{table}[htbp]
    \centering
    \small
    \caption{Label efficiency comparison at $R^2$ thresholds (MM$=$6), showing the number of labeled samples required by random sampling and active learning to reach each threshold.}
    \label{tab:label_efficiency}
    \renewcommand{\arraystretch}{1.0}
    \begin{tabular}{cccc}
    \toprule
    \textbf{$R^2$ Threshold} & \textbf{Random Sampling} & \textbf{Active Learning} & \textbf{Efficiency Ratio} \\
    \midrule
    0.90 & 10,000 & 8,000  & 1.3$\times$ \\
    0.92 & 10,000 & 10,000 & 1.0$\times$ \\
    0.94 & 20,000 & 12,000 & 1.7$\times$ \\
    0.95 & 20,000 & 14,000 & 1.4$\times$ \\
    0.96 & 30,000 & 15,000 & 2.0$\times$ \\
    0.97 & 40,000 & 20,000 & 2.0$\times$ \\
    0.98 & Not reached & 76,000 & --- \\
    \bottomrule
    \end{tabular}
\end{table}

Taken together, the results across all four experimental phases present a coherent picture. CoNN-DAE exhibits strong data efficiency under partial inverse design conditions, as evidenced by performance saturation at 100,000 samples in the upper-bound benchmark, and this efficiency is further enhanced through active learning with informed sample selection. With only 20,000 actively selected labels (2.3\% of the training pool), CoNN-AL already achieves $R^2$ values of 0.967--0.982 across all MM levels, approaching upper-bound performance, and it reaches the $R^2 \geq 0.95$ threshold with only 11,000--14,000 labels, 30--40\% fewer than random sampling at the more difficult MM levels. At the most efficient checkpoint of 76,000 labels, active learning attains $R^2 = 0.9838$ and $\mathrm{MSE} = 0.0164$ at MM$=$6, yielding slightly higher $R^2$ and substantially lower MSE than random sampling at 60,000, UB 100,000, and UB 886,215, although the $R^2$ differences remain small in absolute magnitude. While MAE shows a more mixed pattern, the overall results indicate that targeted sample selection can match or modestly improve upon larger randomly composed training sets using fewer labeled samples. Compared with random sampling under the most challenging scenario (MM$=$6), active learning reduces the labeling requirement by approximately 50\% to reach $R^2 = 0.97$ and is the only approach in this study to reach $R^2 = 0.98$. In the GRC engineering context, where each labeled sample requires CATIA-based modeling and MSC Marc finite element simulation, and assuming simulation cost scales approximately with the number of labeled samples, this corresponds to a potential reduction in simulation cost of about 91\% at 76,000 labels and about 98\% at 20,000 labels relative to the full training pool of 886,215 samples. These results highlight the practical value of integrating active learning with CoNN-DAE for efficient and scalable data-driven partial inverse design.

\section{Conclusion}
\label{sec:conclusion}

This study proposed CoNN-AL, a framework that integrates stream-based active learning with the Cooperative Neural Network for partial inverse design where obtaining labels involves costly simulation procedures. In this framework, Monte Carlo dropout is applied over the cooperative surrogate to estimate predictive uncertainty, which serves as the acquisition signal for real-time query-or-discard decisions as unlabeled candidate samples arrive sequentially from a continuous data stream. By concentrating the labeling budget on the most informative regions of the design space, CoNN-AL reduces the number of costly finite element simulations required to build an effective training set, addressing a key practical bottleneck in data-driven inverse design for engineering applications.

The framework was validated on a real-world automotive glass run channel dataset containing more than 900,000 samples, each generated through CATIA-based parametric design and MSC Marc finite element analysis. The experimental evaluation, structured across upper-bound benchmarking, random sampling baseline, stream-based active learning, and label efficiency analysis, yielded several key findings. First, the upper-bound experiments confirmed that CoNN-DAE achieves performance saturation at 100,000 labeled samples, with negligible improvement when training on the full pool of 886,215 samples, demonstrating the inherent data efficiency of the cooperative architecture. Second, the random sampling baseline revealed a performance plateau around 40,000 samples with diminishing returns thereafter, motivating the need for informed sample selection. Third, stream-based active learning with uncertainty-guided acquisition achieved $R^2$ values of 0.967--0.982 across all missing-mask levels using only 20,000 labeled samples, approximately 2.3\% of the full training pool, approaching the upper-bound performance established with substantially larger datasets. Moreover, active learning reached $R^2 \geq 0.95$ with only 11,000--14,000 labels, approximately 30--40\% fewer than the 20,000 randomly sampled labels required at the more difficult missing-mask levels. Fourth, the label efficiency analysis under the most challenging scenario (MM$=$6) showed that active learning requires up to half the labeled samples of random sampling to reach equivalent $R^2$ thresholds and is the only strategy in this study to attain $R^2 = 0.98$, a level not reached by random sampling within 100,000 labels. At the most efficient active learning checkpoint of 76,000 labels, the model achieved an MSE of 0.0164, approximately 37\% lower than the best random sampling result obtained at 60,000 labels, while also surpassing the upper-bound models trained on 100,000 and 886,215 samples in terms of MSE, although its MAE remained slightly higher than that of the compared models, as discussed in Section~\ref{sec:results_efficiency}. Notably, these results were obtained under the most challenging scenario, where up to half of the design variables are unknown, confirming the effectiveness of the acquisition strategy even under severe design incompleteness.

From a practical standpoint, the results suggest that CoNN-AL can substantially reduce the simulation cost of building training datasets for partial inverse design. In the GRC engineering context, where each labeled sample requires CATIA modeling and MSC Marc finite element evaluation, achieving near upper-bound performance with 20,000 actively selected samples corresponds to a potential reduction of approximately 98\% in simulation effort relative to exhaustive labeling. This level of data efficiency makes data-driven partial inverse design more feasible in industrial settings where simulation budgets are constrained. Together with this study, the GRC dataset is publicly released as a large-scale benchmark to support reproducibility and future research on data-driven inverse design.

Despite these contributions, several limitations should be acknowledged. The current study evaluates CoNN-AL exclusively with the CoNN-DAE variant, which employs a deterministic autoencoder for imputation. While this deterministic formulation is well suited for design reproduction and optimization tasks where a clearly defined output is desired, it does not capture the distributional structure of the design space that generative variants could provide. In addition, the active learning strategy relies on MC-Dropout as the sole uncertainty estimator, and there may be cases where MC-Dropout underestimates or misrepresents the true predictive uncertainty. Deep ensembles, which estimate uncertainty from the disagreement among independently trained models, could be a promising solution for uncertainty estimation and also for the surrogate alternative. The current framework also addresses a single-output inverse design problem, where the target is a scalar sealing force, and its behavior under multi-output objectives has not been investigated.

These limitations point to several directions for future research. Implementing generative variants of the cooperative architecture, such as CoNN with variational or Wasserstein autoencoders, would enable stochastic approaches to partial inverse design, where the model can sample diverse candidate solutions rather than producing a single deterministic output, thereby supporting design space exploration and providing engineers with a richer set of feasible alternatives. Adopting deep ensembles as an alternative uncertainty estimation approach, or as a surrogate architecture, may also improve the quality of the acquisition signal and further enhance label efficiency. Extending CoNN-AL to multi-output inverse design problems, where multiple performance criteria must be satisfied simultaneously, would broaden its applicability to more complex engineering scenarios. Additionally, validating CoNN-AL on engineering problems from other domains, such as structural optimization, materials design, or manufacturing process optimization, would help establish the generality of the approach.

\section*{Declaration of generative AI and AI-assisted technologies in the writing process}
    During the preparation of this work, the authors used OpenAI ChatGPT 5.6 Sol to produce the illustration in Fig.~\ref{fig:grc_dataset}(a), which shows the location of the glass run channel on the vehicle window. After using this tool, the authors reviewed and edited the content as needed and take full responsibility for the content of the published article.

\section*{Acknowledgements}
    This research was supported by Pukyong National University Research Fund. The authors gratefully acknowledge DRB Co., Ltd. for granting permission to publicly release the dataset in standardized form under the Creative Commons Attribution-NonCommercial-ShareAlike 4.0 International License (CC BY-NC-SA 4.0).

%Bibliography
\bibliographystyle{unsrt}
\bibliography{references}

\end{document}